\documentclass{vgtc}                          

\usepackage{mathptmx}
\usepackage{graphicx}
\usepackage{times}

\onlineid{0}

\vgtccategory{Research}

\vgtcinsertpkg
\pdfoutput=1

\title{Procams-Based Cybernetics}

\author{Kosuke Sato\thanks{see: http://www.sens.sys.es.osaka-u.ac.jp/} %
\and Daisuke Iwai$^*$ %
\and Sei Ikeda$^*$ %
\and Noriko Takemura$^*$}
\affiliation{\scriptsize Osaka University}

\teaser{
  \includegraphics[width=.98\hsize]{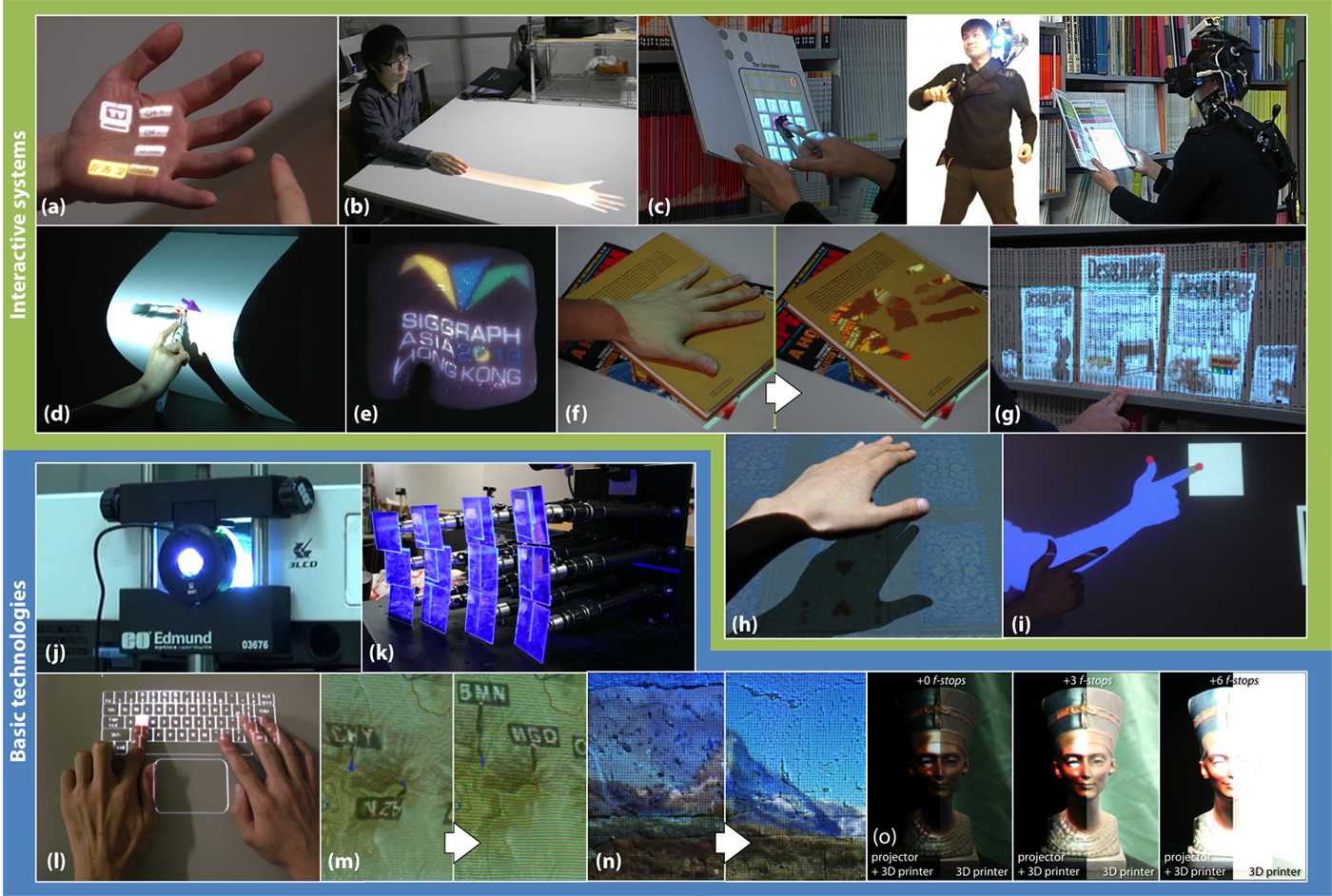}
  \caption{Accomplishments of procams-based cybernetics research: (a) Palm interface \cite{YS07}, (b) extended hand interface \cite{OOIS12}, (c) wearable projection \cite{KS03}, (d) shape deformation interface, (e) deformation visualization, (f) see-through projection and (g) front cover projection for book search support, (h) shadow interface for browsing graphical information on a tabletop  \cite{IIS14}, (i) shadow pointing, (j) Extended depth-of-field (EDOF) projector by fast focus sweep, (k) synthetic aperture projection using a single projector and multiple mirrors for EDOF and shadow-less projection, (l) anywhere touch detection \cite{SIS08}, (m) view management of projected annotations \cite{IYS13}, (n) radiometric compensation \cite{YHS03}, and (o) high dynamic range 3D projection mapping.}
\label{fig:teaser}
}

\begin{document}


\firstsection{Highlights}

\maketitle




Procams-based cybernetics is a unique, emerging research field, which aims at enhancing and supporting our activities by naturally connecting human and computers/machines as a cooperative integrated system via projector-camera systems (procams).
It rests on various research domains such as virtual/augmented reality, computer vision, computer graphics, projection display, human computer interface, human robot interaction and so on.
This laboratory presentation provides a brief history including recent achievements of our procams-based cybernetics project.

\section{Focus}

Procams seamlessly merges computer graphics into real world and allows us to manipulate them without physically constraining our heads and bodies while minimizing renovation of our living spaces.
Leveraging this unique property, our research group focuses on investigating interactive systems to support our daily activities, collaborative works, and machine manipulations (Fig. \ref{fig:teaser}).

\section{Strategy}

To achieve the goal mentioned above, we develop basic technologies to overcome the technical limitations of current projection displays and to achieve robust user manipulation measurement under dynamic projection illuminations to realize flexible and robust interactive systems.

We use cameras, ranging from normal RGB cameras to near and far infrared (IR) cameras, to measure user manipulation as well as scene geometry and reflectance properties.
For the user measurement, we apply IR cameras to robustly detect user's touch actions even under projection-based dynamic illumination.
In particular, we propose two touch detection techniques; one measures finger nail color change by touch action using a near IR camera \cite{SIS08}, while the other measures residual heat of user's touch on a surface using a far IR camera.
For the scene measurement, we use projectors to project either spatial binary or uniform color patterns, and capture the projected results to measure scene depth \cite{SI87} or reflectance properties.
These geometric and photometric information is then used for radiometric compensation that neutralizes the texture of projection surface so that desired color is displayed on textured surfaces \cite{YHS03}.

The followings are basic technologies on projection side, which are developed to change the appearance of real surfaces as freely as computer graphics do.
In contrast, we realized high dynamic range projection systems by boosting the contrast of projection surface texture on relatively flat printed paper of E-ink display, on full color 3D printer output, and on surfaces with dynamic reflectance distributions using photochromic ink that allows us to modulate surface reflectance spatiotemporally \cite{ITHS14}.
We also proposed to extend the DOF of projector by adopting fast focal sweep technique using electrically tunable liquid lens.
Extended DOF projection was realized with another approach, synthetic aperture projection, by which cast shadows could be removed as well.
In order to realize wider field of view projection, we proposed pan-tilt projection and optimized the path of projected image to maximize the spatial resolution.
For displaying legible text information, we proposed an optimization method for projected annotation layout \cite{IYS13}.

\section{Uniqueness}
 
Our research activity is unique since it covers from basic technologies to end-user application systems and lasts for almost 30 years.
It is particularly worth noting that we conducted pioneering and important works that have significantly influenced subsequent researchers.
In 1987, we proposed gray-code pattern projection for shape measurement in active stereo systems \cite{SI87}.
In 2003, we proposed radiometric compensation technique \cite{YHS03} and wearable projection system \cite{KS03}.
In 2007, we introduced palm interface where projector displays graphical information on a user's palm, with which she interacts by touching it \cite{YS07}.


\section{Recent Accomplishments}

We developed several interactive systems to demonstrate that procams-based cybernetics can support interactions between human and computer, human and human, and human and machine, respectively.

For human-computer interaction, we have investigated the following three topics: computing anywhere, shape design support, and book search support.
Aiming at realizing the concept of computing anywhere, we developed wearable projection system in which a user can interact with graphical information projected onto nearby surfaces including her palm by touching it \cite{KS03,YS07}.
To support shape design process, we proposed a system in which a user can manipulate the shape of a real surface, which is visually deformed by projected imagery.
The visualization of a real surface deformation by projection was also proposed.
For book search support, we developed systems that allow users to search for their real books as they do for digital documents stored in their computers.

For human-human communication, we developed an interactive tabletop system in which users can browse their private information by casting their shadows on the tabletop while sharing graphical information in non-shadow areas \cite{IIS14}.
We also proposed shadow-based pointing interface for shared information displayed on a large screen around 10-feet away from users.

For human-machine interaction, we proposed an extended hand interface in which a user's hand is visually extended by projection so that she can manipulate distant appliances as she touch them \cite{OOIS12}.
While seating on a chair, the user can use the extended hand to tell a home service robot a specific distant object such as a cup that she wants the robot to bring or a specific place where the robot should clean up.
We also proposed to enrich the face expression of an android by projecting high frequency face appearance information to realize more natural human-robot interaction.

\section{Future Direction}

\begin{figure}[t]
	\centering
	\includegraphics[width=0.98\hsize]{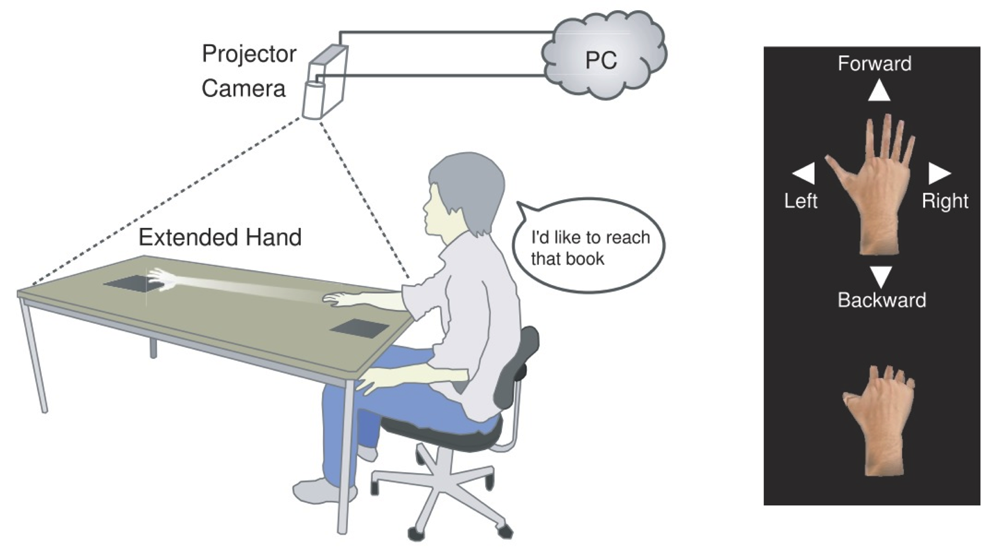}
 \caption{Overview of extended hand interface \cite{OOIS12}.}
 \label{fig:extended}
\end{figure} 

In the extended hand research \cite{OOIS12} (Fig. \ref{fig:extended}), we found an interesting phenomena: a user felt the projected hand is her own hand.
In another research, we found that warmth judgement of an touched object is affected by hand color, which is changed by projection.
These results indicate that we can alter human perception either by projecting user's body texture onto real surfaces or by projecting graphics onto a user's body.
In future, we will focus more on investigating this issue, which could allow us to develop more flexible and intuitive systems, with researchers in other domains such as cognitive science and neuroscience.

\bibliographystyle{abbrv}
\bibliography{paper}
\end{document}